# Using Genetic Algorithm To Evolve Cellular Automata In Performing Edge Detection Edge Detection

Karan Nayak, *MEng Student, Concordia University*

**Abstract**—Cellular automata are discrete and computational models thatcan be shown as general models of complexity. They are used in varied applications to derive the generalized behavior of the presented model. In this paper we have took one such application. We have made an effort to perform edge detection on an image using genetic alogorithm. The purpose and the intention here is to analyze the capability and performance of the suggested genetic algorithm. Genetic algorithms are used to depict or obtain a general solution of given problem. Using this feature of GA we have tried to evolve the cellular automata and shown that how with time it converges to the desired results.

—————————— ◆ ——————————

## 1 INTRODUCTION

Evolutionary computing and genetic algorithms have proven their worth in optimization, serch problems and generalization, in varous applications and fields. They are biologically inspired algorithms relying on processes like population initialization, fitness evaluation, parent selection , applying variation operator , offspring pool and last the survivor selection for next generation . Each process in GA plays a vital role in achieving the desired result.

We will use above said characteristics of the genetic algorithm to evolve the cellular automata which have a fea-ture of self-organizaing themselves using interactions with local neighbourhood. The next state of a cell in CA depends on the state of it's neighbourhood cells and some predefined rule sets. These rule sets govern the organization of the cells in CA.

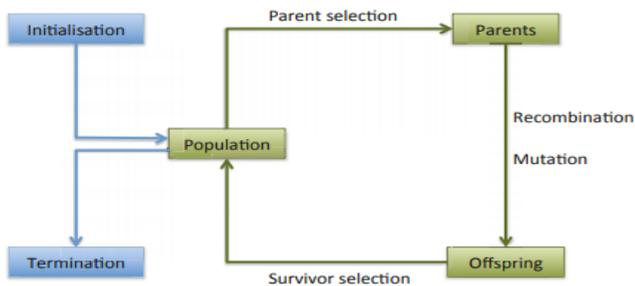

Fig. 1. Flow chart of a Genetic Algorithm.

The application we have picked for our research is edge detection of an image. Edge detection of an image is basically detecting the points or pixels in the image where the image intensity or brightness changes sharply. There aar various mathematical models suggested to perform edge detection.

Below we will use genetic algorithm to drive the evolution of cellular automata in order to realise the edges of a given image. To perfrom edge detection, we need a greyscale image of the given image . Here we have used python programming language to code the suggested algorithm and have made use of some of its standard packages such as OpenCV, Numpy , Json , Matplotlib to better facilitate the realization of the results. We achieved satisfactory results on applying the above said strategies.

Fig. 2. (a) Given Image, (b) Goal Image to be achieved using GA

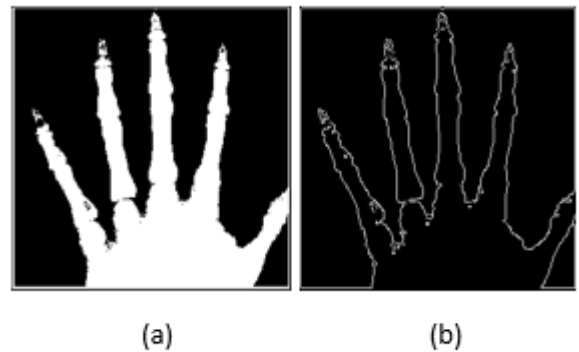

The given image is first converted into greyscal of values 255 and 0. Fig. 1(a) represents that . Then using a OpenCV pyhon canny function the edges of the greyscale image are detected which is shown Fig. 1(b). This image is used as the goal image in the suggested genetic algorithm to achieve the resulting image. The goal_image is a slightly distorted version of the initial image.

————————————————

• Karan Nayak is a student at the Concordia University, Montreal.
E-mail: karannayak07@gmail.com

## 2 CELLULAR AUTOMATA

Cellular automata is a collection of cells on a grid of given dimensions, be it in 2D or 3D. Based on a set of rules, each cell in the CA adjusts its next state. These rules are defiend randomly. CA were studied by a various people and so there are different versions of them like Wolfram's Elementary CA,totalistic CA and Conway's Game of Life CA . Each of them have a different way of defining the rules.In this paper we have taken into account the totalistic cellular automata where the next state of the cell depends on it's surrounding neighbors. This neighbourhood can vary in size. Depending on the neighbourhood one takes into account, accordingly the state of the cell changes. Varous neighbourhoods are shown below.

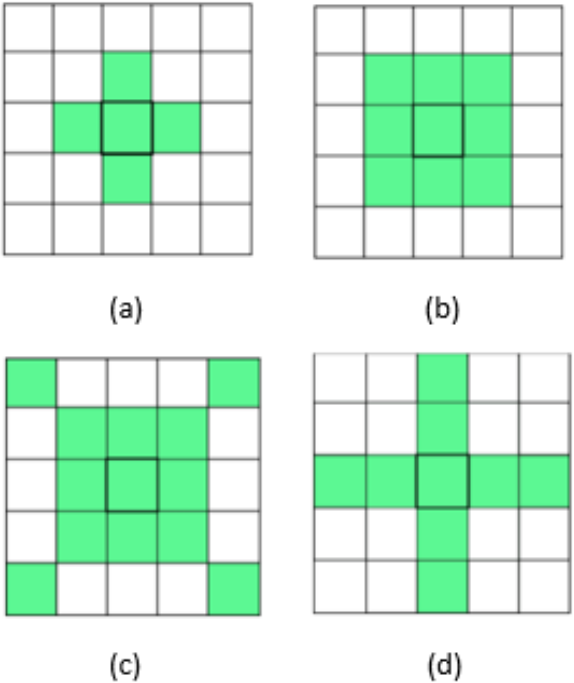

Fig. 3. (a) Von Neumann neighbourhood, (b) Moore neighbourhood, (c) Moore neighbourhood with ears, (d) 9-cell cross

Here , in our suggested approach we have taken the Moore neighbourhood into consideration having a 3x3 window and the middle cell is the core cell whose value will change. The value to be assigned to the cell is defined by the rule set. Rule sets can also be defined in different ways depending on the objective. Here , we have rules defined in binary , that means , there are only two states a cell can assume , either '0' or '1' . The way this works is:-

- As the window is of size 3x3, this window is converted to a 9 bit binary number and the corresponding rule is looked up in the rules table.
- There are $2^9$ = 512 possible rules to lookup for.
- As the window slides over the start image array, it makes a 9 bit binary number and looks up its corresponding rule in the table according to which the state of the core cell is changed.

## 3 GENETIC ALGORITHM

Here, we will discuss how we have evolved the CA using genetic alogorithm, the strategies used at different stages in GA and the results.

TABLE 1
PSEUDO CODE FOR FITNESS EVALUATION

| Pseudo Code: |
| --- |
| For g in generations:<br>    For i in range(len(ind_pool)):<br>        For p in passes: #number of passes on the array<br>            Result_array = slide_window(start_array)<br>            Cal_fitness(result_array , goal_image)<br><br>    Parents = Elite + Random # 2 elite and 2 random<br><br>#perform crossover<br><br>    For j in Parents:<br>        Select two parents randomly<br>        o1, o2 = xover2(p1 , p2)<br>    Fitness_evaluator(offspring_pool) # slides window over the start image with reference to new offsprings.<br><br>#perform mutation<br>    For I in offspring_pool:<br>        Mutation1(o[i])<br>    Fitness_evaluator(offspring_pool)<br><br>    Survivor_select = offspring_pool + ind_pool<br>    Survivor pool = (20% best of survivor_select)  +<br>                (80% random of survivor_select)<br><br>    ind_pool = survivor_pool #passing on to next gen |

### 3.1 Population and Representation

To start of, the first thing is to define your population that will undergo the algorithm process. We will call each rule set or table as an individual in terms of GA. Based on the population size given our code will generate random rule set for each individual. Here, each rule in the rule set acts as a gene that can be altered using variation operators. Next important thing is how to represent these individuals , binary , integers , float , complex numbers . Here , as our image is in binary form of '1' and '0' , the window is also converted to binary form of 9-bits and the rule states are also defined in binary , so we have represented each individual our population as list of binary rules , each list representing an individual , containing of 512 rules. The population is stored as a key:value pair in a json file .

TABLE 2
INDIVIDUAL IN A POPULATION

| Population JSON | |
| --- | --- |
| 9 bit Binary window rule | Rule state |
| '000000000' | 0 |
| '000000001' | 1 |
| '000000010' | 1 |
| . | . |
| . | . |
| . | . |
| '111111111' | 0 |

## 3.2 Fitness Evaluation and Parent Selection

Fitness evaluation is doen to select the parents that will undergo further process of the algorithm. Fitness evaluation works as a filter to drain out unwanted individuals that are not driving the algorithm towards the solution. It Fitness evaluation tells how good an individual is to be passed on for further processing. In edge detection, fitness can be seen as similarity between the result image and the goal image. This can be achieved using RMSE measure, hamming distance measure and various other measures. In our paper, we are using Hamming Distance to calculate the fitness of an individual. Each individual iterated over the start_image array and the resulting image is compared with the goal image, giving us the fitness , measure of that individual. So a low hamming distance indiciates a good fitness. Fitness evaluation is done at multiple stages of the GA.

TABLE 3
PSEUDO CODE FOR FITNESS EVALUATION

| Pseudo Code: |
|---|
| For i in rows of goal_image:<br>    For j in column of goal_image:<br>        If start_image_pixel != goal_image_pixel<br>        Fitness += Fitness |

Parent selection is done based on the fitnesses of the indviduals. In ot approach we have incorporated 20% elitism while selecting the parents which ensures that best individuals get passed on for further processing, hence driving the GA towards the desired solution. However it is important to have diversity in parent pool otherwise chances of premature convergence increases and the GA might get stuck in local optima. So we are also selecting 20% individuals randomly.

$$Parent\ pool = Elite\ parents + random\ parents \quad (1)$$

## 3.3 Recombination

Recombination is a biological term which involved two parents mating and producing the corresponding offsprings. These offsprings will have genes of both the parents. Recombination can change the fitness of an individual drastically depending on the strategy being used. The most common one is n-point crossover where an individual is dissect at 'n' random points and those sections are exchanged between the select parents , giving us two offsprings.

We have incorporated two types of recombination strategies, 1-point crossover and 2-point crossover, to show the performance of the GA . The results achieved are shown below :-

We can observe that with the passing of each generationcitation the fitness of the best individual decreases. Recombination doesn't always guarantee improvement as some genes of the parents might weaken the fitness of the individual.

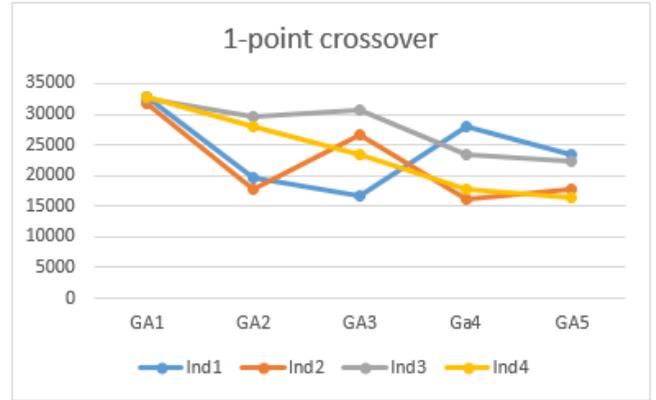

(a)

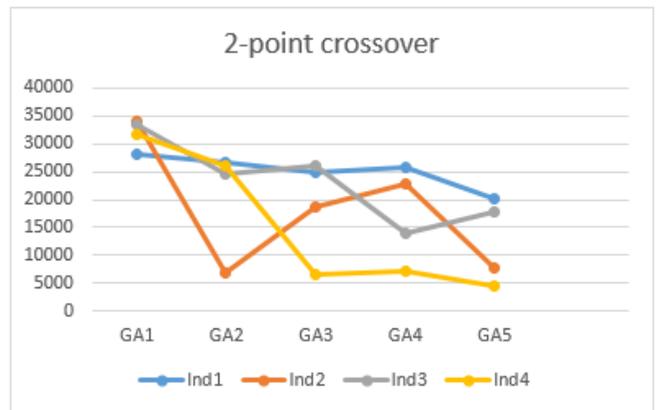

(b)

Fig. 4. (a) 1-point crossover (b) 2-point crossover

## 3.4 Mutation

Mutation requires only one parent to produce the offspring. For binary representation mutation strategies like gene alteration , swap mutation and other can be applied . In our approach we have taken into account two mutation strategies.

### 3.4.1 Mutation Type 1

After running our algorithm couple of time we observed that our best solution was often getting stuck in the local optima , which was happening if the candidate got mutated too much in previous generation. In our case , if the individual was having more '0' in it's genes then it was overconverging the solution. So to tackle this, we were altering the genes in the individual based on the fitness of the individual. The algorithm is designed in a way that , it will alter more genes from '1' to '0' if the fitness is bad and it will alter less genes if the fitness is less and going in the direction of solution. Also if the fitness is stuck in local optima then it will mutate the selected genes with '1'. So that the individual can get a push to move forward or backward.

TABLE 4
PSEUDO CODE FOR MUTATION TYPE 1

| Pseudo Code: |
|---|
| If fitness stuck in local optima: |
|    Select few genes based on fitness: |
|    Alter all those genes = 1 |
| Else: |
|    Mutation_rate = 0.5 |
|    Generate random float between (0,1) |
|    If float < 0.5: |
|       Gene = 0 |
|    Else: |
|       Continue |

### 3.4.1 Mutation Type 2

As suggested in [1], any rule set contains some basic rules using which you can derive other rules. Here we have $2^9$ rules out of which 9 rules are the basic ones. These basic rules are [1,2,4,8,16,32,64,128,256,512]. Additon of any random number from the given list will give us another linear rule. In our approach we are settting the genes of the given numbers in the list to '0' and also few other genes derived from them.

TABLE 5
PSEUDO CODE FOR MUTATION TYPE 2

| Pseudo Code: |
|---|
| Mutation_rate = 0.5 |
| Generate random float between (0,1) |
| If float < 0.5: |
|    For i in range(9): |
|       Val = 2**i |
|       If pool[Val] == 1: #pool is the individual |
|          Pool[Val] = 0 |
|       Else: |
|          Continue |
|    For i in range(4): #run the loop four times |
|       k = pick two random no. from the above list |
|       temp = k1 + k2 |
|       If pool[temp] == 1 |
|          pool[temp] = 0 |
|       Else: |
|          Continue |
|    Return pool |
| Else: |
|    Return pool |

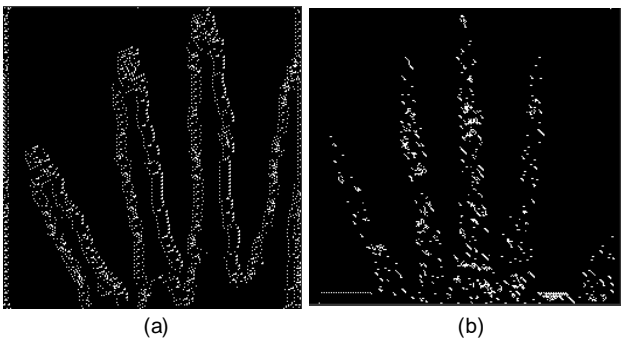

Fig. 5.(a) Mutation type 1 result (b) Mutation type 2 result

Here in mutation type 2 we observed that the solution was generating results but not upto the mark.

## 4 SURVIVOR SELECTION

Survivor pool depicts the individuals that will be passed on to the next generation. Here also we are maintaining the elitism where in we are selecting 20% best individuals and rest 80% randonmly to maintain the diversity in the next generation.

## 5 RESUTLS

Good results in our approach were achieved by using 2-point crossover and mutation type 1. Below observations were made from the performance :-

- Based on the dimensions of the image array, the execution time of one generation varies. For an array of 256x256, one generation takes approximately 56.1 seconds.Majority of the time was spent in sliding the window through the array. Larger the array, more time it took.
- The GA was giving best results when ran for 5 generations, after which it was overconverging the result.
- The avg fitness was improving with the passing pf each generation.

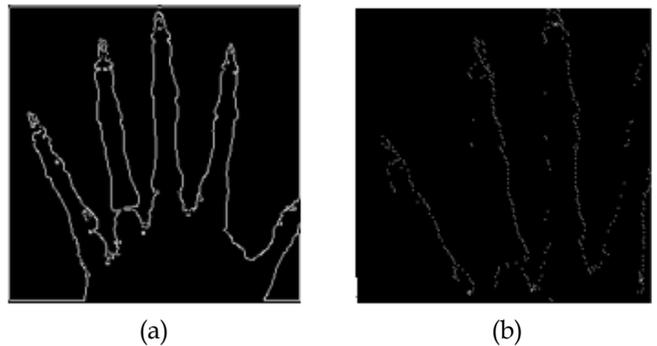

Fig. 6 (a) Goal image (b) result image .

Below show are some of the graphs to viualize the performance of the GA.

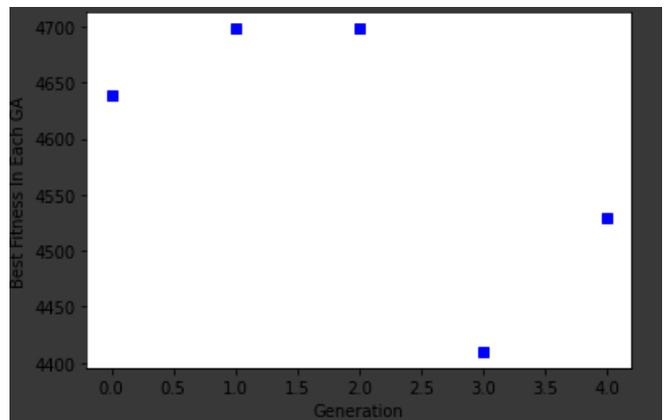

Fig. 7.Graph of best fitness in each generation.

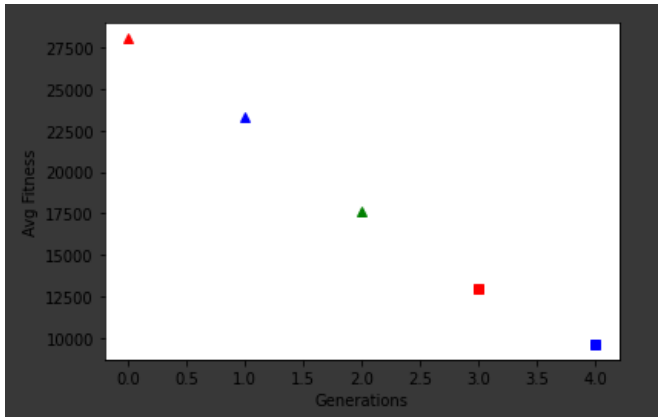

Fig. 8.Average fitness in each generation.

## 6 EXPERIMENTS

We also tried the same GA on different for different images of different dimensions and below were the results.

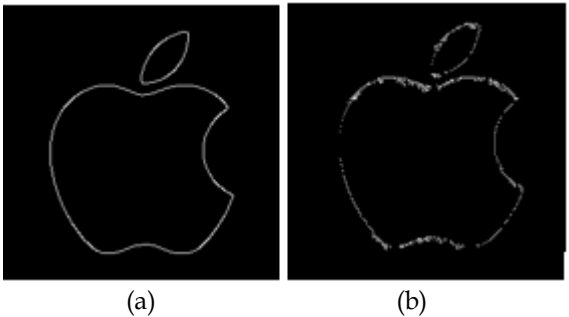

       (a)                  (b)

Fig. 9.(a) Goal Image (b) Result Image

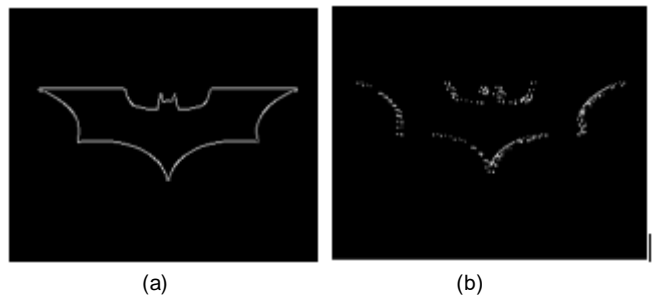

       (a)                  (b)

Fig. 9.(a) Goal Image (b) Result Image

## 7 CONCLUSION

In this paper we have made an effort to evole a cellular automata using genetic algorithm to perform an edge detection of a given image. We observed and analuzed different parameters of GA and saw how vital they are in realizing the correct solution.